\documentclass{article}
\usepackage[utf8]{inputenc}

\usepackage[T1]{fontenc}
\usepackage{times}%

\usepackage{amsmath}
\usepackage{dcolumn}
\usepackage{graphics}

\usepackage{graphicx}
\usepackage{color}
\usepackage{booktabs}
\usepackage{multirow}
\usepackage{enumitem}  
\usepackage{hyperref}

\usepackage{authblk}
\usepackage{blindtext}

\title{TxPI-u: A Resource for Personality Identification of Undergraduates\footnote{Preprint of \cite{ramirezEtal2018}. The final publication is available at IOS Press through \url{ https://content.iospress.com/articles/journal-of-intelligent-and-fuzzy-systems/ifs169484}}}

\author[1]{Gabriela Ramírez-de-la-Rosa\thanks{Corresponding author.}}
\author[1]{Esaú Villatoro-Tello}
\author[1]{Héctor Jiménez-Salazar}
\affil[1]{Language and Reasoning Research Group, Information Technologies Department, Universidad Aut\'onoma Metropolitana (UAM) Unidad Cuajimalpa, M\'exico}

\affil[ ]{\textit {\{gramirez,evillatoro,hjimenez\}@correo.cua.uam.mx}}

\date{May 2018}

\begin{document}

\maketitle

\begin{abstract}
Resources such as labeled corpora are necessary to train automatic models within the natural language processing (NLP) field. Historically, a large number of resources regarding a broad number of problems are available mostly in English. One of such problems is known as Personality Identification where based on a psychological model (e.g. The Big Five Model), the goal is to find the traits of a subject's personality given, for instance, a text written by the same subject. In this paper we introduce a new corpus in Spanish called Texts for Personality Identification (TxPI). This corpus will help to develop models to automatically assign a personality trait to an author of a text document. Our corpus, TxPI-u, contains information of 416 Mexican undergraduate students with some demographics information such as, age, gender, and the academic program they are enrolled. Finally, as an additional contribution, we present a set of baselines to provide a comparison scheme for further research.

\end{abstract}


\section{Introduction}\label{sec:intro}
There is a growing interest in the computer science community on studying individual's personality. This new interest, mainly among natural language processing community is due to the fact that through traditional techniques developed by psychologists, identification of one's personality has been proved efficient for predicting thought patterns, emotions and behaviour \cite{Funder2001personality}. Particularly, knowing this kind of information from an individual has been useful to detect her/his well-being as well as it is been helpful in the study of her/his mental health. For instance, in the past,  knowing the personality of a person allowed physicians to prevent mental disorders or mental conditions \cite{Ozer2006Personality}. 

Automatically identifying a person's personality is a relevant task in several areas of Computational Sciences. As an example, in the Human-Computer Interaction field, on one hand, knowing the personality of an user can help to improve its experience with the system; on another hand, providing with a compatible personality to the system itself may result in a more natural interaction with the final user \cite{Bickmore2005}. For instance, in video games, the notion of a character's personality has been a key factor to improve characters' credibility \cite{Andre1999Integrating}; in education, automated tutors could be more effective in reaching students if the tutor adapts to the student's personality \cite{Komarraju2005}.

In order to build systems like the ones in previous examples, it is necessary to have resources, i.e. labelled corpora, to be used in building automatic systems to effectively identify a person's personality. Historically, a large number of resources regarding a broad number of problems are available mostly in English. Particularly, for personality identification task there are very few resources  available (English included) in comparison to other older problems. 

This lack of resources difficult the development of such systems. In some cases, researchers have to build their own dataset which is an expensive task, in time, effort and money. Whilst there have been some attempts to build public available resources, to the best of our knowledge, there are no such resources for Spanish. 

In order to overcome this obstacle and to help advance forward the research in this growing area we introduce a resource of Spanish writing texts of 416 Mexican undergraduates students. Each text is accompanied with the personality traits obtained with a psychological instrument called TIPI (Ten Item Personality Inventory); additionally, each text is also accompanied with the gender and age of the author. This extra information can be also useful for researchers investigating problems such as author profiling in Spanish texts.

The rest of this paper is organized as follows. First, in Section \ref{sec:bfmodel} the Big Five Personality Model (BF) is described to illustrate what is been measured to each individual participating in our study. Section \ref{sec:relatedwork} shows some related work to the current datasets for personality identification. Section \ref{sec:makingcorpus} presents step by step how the corpus proposed was compiled, as well as some general information about it. After that, a statistical analysis of our corpus can be found in Section \ref{sec:description}. Once we know the gist of our corpus, Section \ref{sec:comparison} shows some similarities between our introduced corpus against a bigger corpus of essays written in English, this English corpus is one of the most used in the NLP community. As an additional contribution, in Section \ref{sec:baselines} a set of baselines is presented using TxPI-u corpus. And finally, in Section \ref{sec:conclusions} conclusions and perspectives are presented.

%%%%%%%%%%%%%%%%%%%%%%%%%%%%%%%%%%%%%%%%%%%%%%%%%%%%%%%%%%%%%%%%%%
%%%%%%%%%%%%%%%%%%%%%%%%%%%%%%%%%%%%%%%%%%%%%%%%%%%%%%%%%%%%%%%%%%
%%%%%%%%%%%%%%%%%%%%%%%%%%%%%%%%%%%%%%%%%%%%%%%%%%%%%%%%%%%%%%%%%%

\section{Big five personality model} \label{sec:bfmodel}

The individual's personality is determined by her or his stable patterns of behavior shown in any particular situation. In other words, the personality is defined by those characteristics that do not change, and that are independent of the situation in which a person is involved \cite{VinciarelliSurvey2014}.

Goldberg established that psychological models based on traits are more efficient for measuring aspects in the life's subject \cite{Goldberg1993Structure}. Such personality traits are internal dispositions that exhibit processes such as thinking, feeling or acting in specific situations resulting in the same result \cite{Wrzus2015}.

The dominant model based on traits is known as the Big Five Model (BF) or Five-Factor Model (FFM) \cite{McCrae2002}. This model proposes five traits with two poles each, positive and negative. The descriptions of these traits are as follows:

\begin{itemize}
    \item \textit{Extroversion} is associated with energy, positive emotions, assertively, sociability and expressively; its negative pole is \textit{introversion}.
    \item \textit{Emotional stability} is associated with controlling impulses, its negative pole is \textit{neuroticism} which is the tendency to experience unpleasant emotions such as angry, anxiety, depression or vulnerability.
    \item \textit{Agreeableness} refers to the tendency to be understanding and cooperative. Its negative pole refers to distrust and apathy towards others.
    \item \emph{Conscientiousness} is the tendency to show auto-discipline, to act in a loyal way, to reach goals and to plan, to be organize and trustworthy. Its negative pole refers to spontaneous behaviors.
    \item \emph{Openness to experience} is associated with appreciation of unusual ideas, and with imaginative and curious minds. The negative pole of this trait is associated with being unimaginative and inflexible to change.
\end{itemize}

Traditionally, to identify to what extend each trait is present in one individual, psychologists have developed standard questionnaires. The more frequently used questionnaires are: NEO-Personality-Inventory Revised (NEO-PI-R with 240 questions) \cite{CostaMcCrea1992} and the Big-Five Inventory (BFI with 44 questions) \cite{john1991big}. In this study we used the Ten Item Personality Inventory (TIPI with only 10 questions) \cite{Gosling2003}.

%%%%%%%%%%%%%%%%%%%%%%%%%%%%%%%%%%%%%%%%%%%%%%%%%%%%%%%%%%%%%%%%%%
%%%%%%%%%%%%%%%%%%%%%%%%%%%%%%%%%%%%%%%%%%%%%%%%%%%%%%%%%%%%%%%%%%
%%%%%%%%%%%%%%%%%%%%%%%%%%%%%%%%%%%%%%%%%%%%%%%%%%%%%%%%%%%%%%%%%%

\section{Related work} \label{sec:relatedwork}
According to Pennebaker, language is a good indicator of our personality, that is because through language we can express our way of thinking and feeling \cite{Pennebaker2011Secret}. Consequently, there is a great amount of studies focusing on the analysis of expressions of language, such as those present in texts produced by a person. One of the first works in this area was based entirely on identifying types of words used in such texts \cite{Argamon2005}. Following this line of research, several other studies have conducted analysis of written texts from blogs or essays \cite{mairesse2007,oberlanderNowson2006,Iacobellietal2011}; or in social media \cite{Farnadietal2016,celli2013relationships,parketal2015}.

While some researchers have gathered their own resources to conduct their investigations, there are few main resources broadly used: i) a corpus collected by Pennebaker \cite{Pennebakeretal99} and Mairesse \cite{mairesse2007} (\textit{Essays corpus}) that consists of 2479 essays from psychology students, ii) \textit{myPersonality corpus}, a collection of Facebook's posts \cite{kosinskiEtal2015}, and iii) the \textit{PAN-AP-15 corpus} developed in the framework of the PAN 2015\footnote{\url{http://pan.webis.de/clef15/pan15-web/author-profiling.html}} author profiling shared task \cite{rangel2015overview}, where Twitter data is provided in several languages including Spanish. These freely available datasets are different in their nature, whilst \textit{Essays corpus} has long texts and has been gathered through several years, \textit{myPersonality} and the \textit{PAN-AP-15} corpora are examples of a massive short-texts data gathered from social media domains.

Although the \textit{PAN-AP-15 corpus} has small partition of tweets in Spanish, our corpus is more directly related to the Essays corpus. As mentioned before, our goal is to contribute in providing resources for promoting studies of personality identification in the Spanish language. In addition, we want to provide a reference study on the performance of existing methods and algorithms in solving the posed task using our corpus. 
In Section \ref{sec:comparison} a more detailed comparison between Essay corpus and TxPI-u corpus is presented.
 
%%%%%%%%%%%%%%%%%%%%%%%%%%%%%%%%%%%%%%%%%%%%%%%%%%%%%%%%%%%%%%%%%%
%%%%%%%%%%%%%%%%%%%%%%%%%%%%%%%%%%%%%%%%%%%%%%%%%%%%%%%%%%%%%%%%%%
%%%%%%%%%%%%%%%%%%%%%%%%%%%%%%%%%%%%%%%%%%%%%%%%%%%%%%%%%%%%%%%%%%

\section{Making the TxPI-u corpus}\label{sec:makingcorpus}

The TxPI-u (Text for Personality Identification of Undergraduates) corpus is a resource that can be used for building automatic systems for personality identification task. This corpus consists of texts written in Spanish from undergraduates Mexican students. In the following we describe the methodology employed to assemble such corpus.

%%%%%%%%%%%%%%%%%%%%%%%%%%%%%%%%%%%%%%%%%%%%%%%%%%%%%%%%%%%%%%%%%%

\subsection{The sample} \label{subsec:sample}

Every year the Autonomous Metropolitan University campus Cuajimalpa (Universidad Autónoma Metropolitana Unidad Cuajimalpa) receives undergraduate students. During 2016, the University received near 600 students for 10 different academic programs. 

During the fourth week of classes we attended to the classrooms to ask for the students' participation in our study. The students were informed about the research we were conducting and 417 decided to collaborate. The distribution of participants (also referred as subjects) per academic program is shown in Table \ref{table:sujetosporcarrera}. 

\begin{table}[tbp]
\centering

\caption{Participants in the corpus divided by gender per academic program in the 2016 admission process at the University} \label{table:sujetosporcarrera}

\begin{footnotesize}
\begin{tabular}{lrrr}
\hline
Academic program &	Male	&	Female	&	Total\\
\hline
    Management	&	6	&	17	&	23	\\
    Humanities 	&	19	&	25	&	44	\\
    Social-territorial Studies	&	12	&	12	&	24	\\
    Communication Sciences	&	30	&	30	&	60	\\
    Design	&	15	&	37	&	52	\\
    Tech. \& Information Systems	&	28	&	16	&	44	\\
    Computational Engineering	&	45	&	15	&	60	\\
    Applied Mathematics	&	14	&	5	&	19	\\
    Biological Engineering	&	20	&	23	&	43	\\
    Molecular Biology	&	19	&	29	&	48	\\
\hline
    Total    &	208	&	209	&	417	\\
\hline
\end{tabular}
\end{footnotesize}

\end{table}

\begin{figure*}[tbp]
\includegraphics[width=1\textwidth]{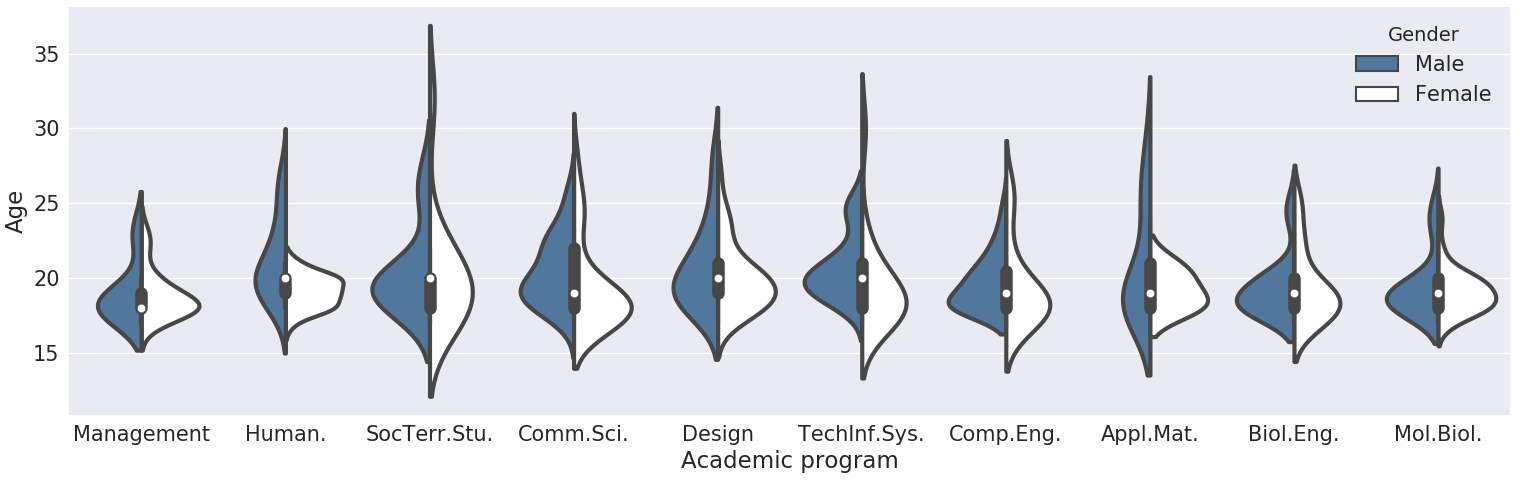}
\caption{Distribution of subjects per age, gender and academic program. Total number of subject is 417, the data for this graph came partially from Table \ref{table:sujetosporcarrera}.} \label{fig:agegenderdegree}
\end{figure*}

As we can see the corpus is balanced in terms of gender. There is also a representation of each field of study from social sciences to mathematics and engineering. Since the subjects were starting the undergraduate education almost everyone is between 19 and 21 years old. The complete distribution of age and gender per academic program can be seen in Figure \ref{fig:agegenderdegree}.

%%%%%%%%%%%%%%%%%%%%%%%%%%%%%%%%%%%%%%%%%%%%%%%%%%%%%%%%%%%%%%%%%%

\subsection{The instrument} \label{subsec:instrument}
The participation in our study consisted in answering an instrument. The goal of such instrument was twofold. First, to determine the subjects' personality in order to label the data according to the Big Five model; and second, to collect a sample of written text from all the subjects regarding personal experiences.

Consequently, the instrument was designed with three parts in order to gather: i) general information, ii) answers for the personality test, and iii) a handwritten short essay.

\begin{enumerate}[label=(\roman*)]
    \item The first section, general information, was designed to get contact information of the subject, also her or his gender, the academic program and her or his social media accounts (such as Facebook and Twitter). It is worth to mention that a small percentage of the subjects indicated a social media account; thus, such information is not included in the final corpus.
    \item The second section has a personality test. In order to take as few time as possible from the participants, the Ten Item Personality Inventory \cite{Gosling2003} in Spanish \cite{RenauEtal2013} was used. 
    \item The last part of the instrument included one instruction and a blank page. The instruction was given in Spanish and can be translated as: \textit{Tell us about yourself, for instance, something about your family's history or an event you think was relevant in your life that comes to your mind}. 
\end{enumerate}

%The Spanish version of this instrument is available in \url{lyr.cua.uam.mx/resources/personality/TxPIu/}.

%%%%%%%%%%%%%%%%%%%%%%%%%%%%%%%%%%%%%%%%%%%%%%%%%%%%%%%%%%%%%%%%%%
%%%%%%%%%%%%%%%%%%%%%%%%%%%%%%%%%%%%%%%%%%%%%%%%%%%%%%%%%%%%%%%%%%
%%%%%%%%%%%%%%%%%%%%%%%%%%%%%%%%%%%%%%%%%%%%%%%%%%%%%%%%%%%%%%%%%%
%%%%%%%%%%%%%%%%%%%%%%%%%%%%%%%%%%%%%%%%%%%%%%%%%%%%%%%%%%%%%%%%%%

\section{Description of TxPI-u corpus} \label{sec:description}
%%%%%%%%%%%%%%%%%%%%%%%%%%%%%%%%%%%%%%%%%%%%%%%%%%%%%%%%%%%%%%%%%%

\subsection{The essays and its transcriptions} \label{subsec:transcription}
To ease the application of the instrument to all students, our instrument was applied on paper. During the digital transcriptions of the handwritten essays we noticed some particularities of handwriting, i.e. small modifications of words, the intent to erase a word, insertion of letters into words, or words into sentences, incorporation of emojis or drawings, as well as misspelling and syllabification. We believe that analyzing such handwriting phenomena would be useful for a better understand on how people with certain trait of personality thinks and behaves. Consequently, the TxPI-u corpus provides two version of the essays, with and without these labels. Although in a digital environment this type of phenomena would be difficult to capture, our intention in labeling such information is to analyze if there is a direct correlation among these and the subjects' personality traits.

Thus, we used seven labels, namely: <FO:well-written word> (misspelling), <D:description> (drawing), <IN> (insertion of a letter into a word), <MD> (modification of a word, that is a correction of a word), <DL> (elimination of a word), <NS> (when two words were written together; e.g. \textit{Iam} instead of \textit{I am}) and, SB (syllabification). An example of such tagging is given in Table \ref{table:extranscript}.

\begin{table*}[tbp]
\centering
\caption{Example of a subject hand written essay in Spanish, its manual transcription with added tags and its corresponding English translation} \label{table:extranscript}
\begin{footnotesize}

\begin{tabular}{p{0.55in} p{4in}}
\hline
 & {\includegraphics[width=3.8in]{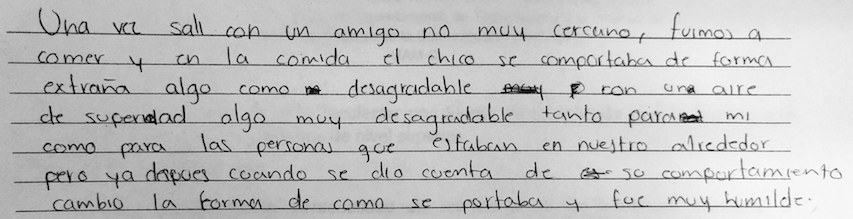}} \\ 
%\hline
Manual transcription & Una vez sali {\color{red}<FO:salí>} con un amigo no muy cercano, fuimos a comer y en la comida el chico se comportaba de forma extraña algo como {\color{blue} <DL>} desagradable {\color{blue} <DL>} {\color{blue} <DL>} con un {\color{magenta} <MD>} aire de superioridad {\color{magenta} <MD>} algo muy desagradable tanto para {\color{blue} <DL>} mi {\color{red} <FO:mí>} como para las personas que estaban en nuestro alrededor pero ya despues {\color{red} <FO:después>} cuando se dio cuenta de {\color{blue} <DL>} su comportamiento cambio {\color{red} <FO:cambió>} la forma de como {\color{red} <FO:cómo>} se portaba y fue muy humilde. \\
%\hline
English translation & Once I went out with a friend not so close to me, we went to eat and while eating the guy was acting a little weird kind of rude as he was superior to me, it was rude for me as for the people around us but after he realized his behavior he changed the way he was acting and he was humble.\\
\hline
\end{tabular}
\end{footnotesize}

\end{table*}

\begin{table}[tbp]%tbph!]%htb]
\centering
\caption{Correlation among tags; where tags FO, D, IN, MD, DL and NS means misspelling, drawing, insertion of some letter, modification of some word, elimination of some word, do not separate two words, and syllabification, respectively} 
\label{table:corrALLtags}
\begin{footnotesize}

\begin{tabular}{rrrrrrr}
\hline
& FO  & D &  IN & MD & DL & NS \\
\hline
%    FO & 1.00 &  &  &  &  &   &   \\
    D & -0.04 & & & & & \\
    IN & -0.01 & -0.01 &  & & &   \\
    MD & 0.00 &  0.01 &  0.00 &  & &    \\
    DL & 0.02 & -0.02 & -0.03 &  0.05 &  &   \\
    NS & 0.08 & -0.03 & -0.03 &  0.07 &  0.06 &    \\
    SB & -0.04 & -0.02 & -0.03 &  0.02 & -0.06 &  0.00 \\
\hline
\end{tabular}
\end{footnotesize}

\end{table}

We compute the Pearson correlation of the percentage of presence in each essay to analyze how these seven tags are correlated among them. Table \ref{table:corrALLtags} shows that the maximum value of correlations is still below 0.1. Nevertheless, the more correlated tags are \textit{NS} and \textit{FO} with 0.08 and also, \textit{NS} and \textit{MD}  with 0.07.

\begin{figure}[tbp]
\centering
\includegraphics[width=0.6\textwidth]{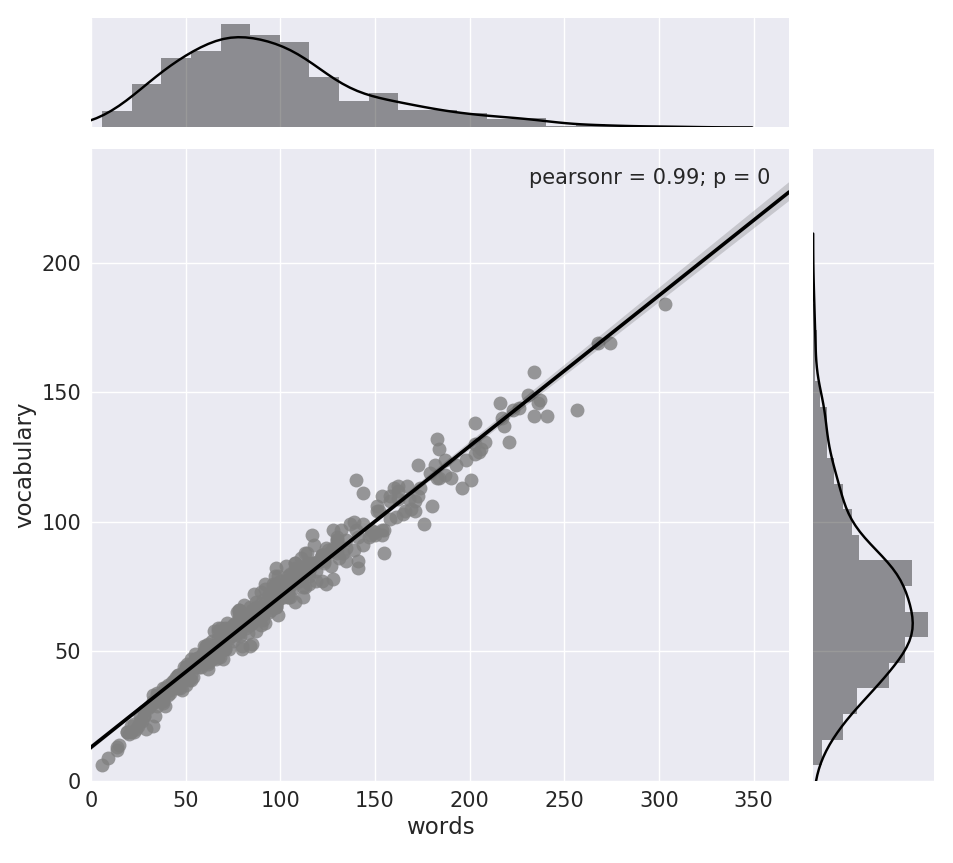}
\caption{Distribution of number of words and total vocabulary per essay and the correlation among these two variables in the TxPI-u corpus. Shown frequency distribution graph (above and right) describes one variable independently of the other.} \label{fig:vocwordsALL}
\end{figure}

We also measured the number of words and total vocabulary used by the analyzed subjects. Figure \ref{fig:vocwordsALL} presents the correlation between these two values and shows, as is expected, that the more written words the greater the vocabulary used. Note that to calculate this measures we did not used any kind of lemmatization tool.

%%%%%%%%%%%%%%%%%%%%%%%%%%%%%%%%%%%%%%%%%%%%%%%%%%%%%%%%%%%%%%%%%%

\subsection{Personality information} \label{subsec:personality}
For the second part of the instrument (see Section \ref{subsec:instrument}) we registered the numeric value computed for each trait according to the answers to the Ten Item Personality Inventory (TIPI) test \cite{Gosling2003}. 

\begin{figure*}[tbp]
\centering
\includegraphics[width=1\textwidth]{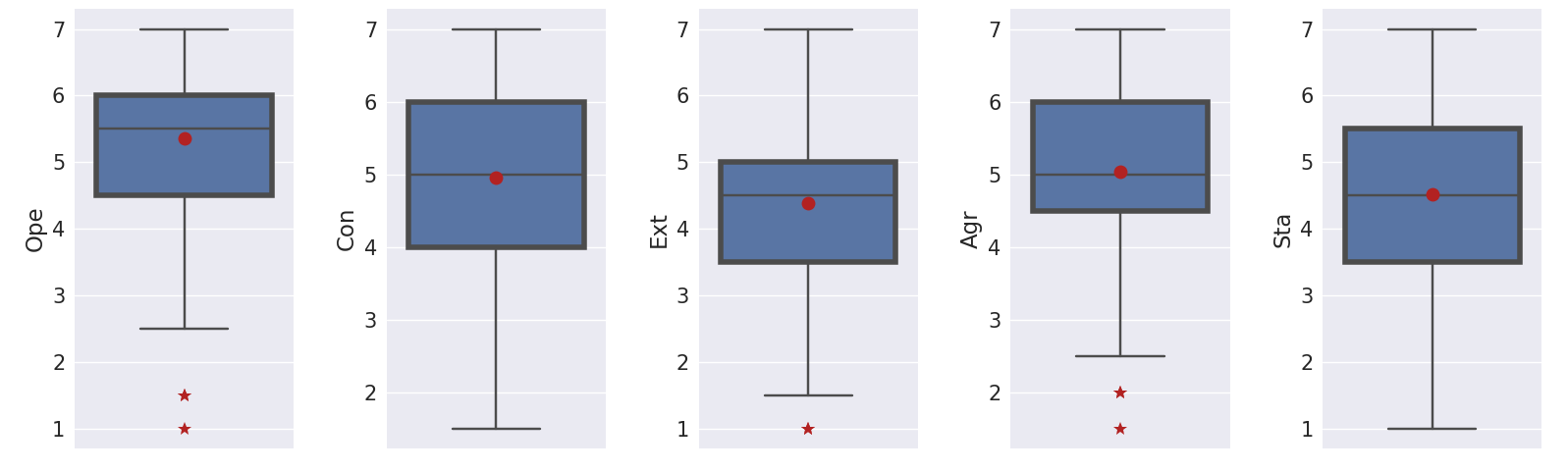}
\caption{Distribution of numerical values of each trait according to the answers of each subject to the TIPI test. The maximum value is 7 and the minimum is 1. The media value is shown inside the box by a circle.} \label{fig:numtipiALL}
\end{figure*}

The TIPI test includes two questions (items) for each of the five traits of the Big Five Model. A more detailed explanation of how to compute the personality of any given answer is presented in \cite{Gosling2003}. Consequently, this test allows to have a numeric value between 1 and 7 to each trait, Figure \ref{fig:numtipiALL} shows the distribution of this numeric value.

To have a general view of numerical values of each trait, the correlation between traits was computed with a Pearson correlation. In Table \ref{table:corrALLnumtraits} all correlations are shown. It can be seen that traits more positively correlated are Emotional Stability and Agreeableness with a value of 0.34. Emotional Stability is also positively correlated to Consciousness (with 0.28). While the correlation is small, we can say that stable subjects are also, to some extend, agreeable and reasonable. Another positive correlation exists between Openness and Extroversion (0.28), indicating that subjects open to new experiences are also, to some extend, extroverts.

\begin{table}[tbp]
\centering
\caption{Correlation values among traits according to the answers to the TIPI test. Ext, Agr, Con, Sta and Ope stand for Extroversion, Agreeableness, Consciousness, Emotional Stability, and Openness, respectively} \label{table:corrALLnumtraits}

\begin{footnotesize}
\begin{tabular}{lrrrrr}
\hline
 &	Ext	& Agr &	Con  &  Sta	& Ope\\
\hline
    Ext & 1.00 &  &  &  &   \\
    Agr & -0.06 &  1.00 &   &   &   \\
    Con & 0.08 &  0.23 &  1.00 &  &   \\
    Sta & 0.09 &  0.34 & 0.28 &  1.00 &  \\
    Ope & 0.27 & 0.15 &  0.16 &  0.07 &  1.00 \\
\hline
\end{tabular}
\end{footnotesize}
\end{table}

Then, according to Gosling's normative data for the TIPI questionnaire, the 417 subjects were classify with five traits. Each of them were labeled into four classes: high, medium high, medium low, and low. Table \ref{table:sujetosporclase} shows the number of subjects per class of each personality trait. Each trait has its own normative values; which were obtained  with 1704 subjects of different ethnicities \cite{Gosling2003}. One subject was removed from the collection since her/his essay was empty; therefore, the total number of subjects in TxPI-u is 416. 

\begin{table}[tbp]
\centering
\caption{Number of subjects per class of each personality trait according to the normative data for the Ten-Item Personality Inventory (TIPI) given by Gosling et al. \cite{Gosling2003}} \label{table:sujetosporclase}
\begin{footnotesize}
\begin{tabular}{lrrrr}
\hline
Trait &	High	& Medium	&	Medium  &  Low	\\
      &         & High      &   Low & \\
\hline
    Openness	&	91	&	145	&	116	& 64 \\
    Consciousness	&	19	&	150	&	138	& 109 \\
    Extroversion	&	72	&	137	&	169	& 38 \\
    Agreeableness	&	60	&	115	&	151	& 90 \\
    Emotional Stability	&	34	&	151	&	151	& 80 \\
\hline
\end{tabular}

\end{footnotesize}

\end{table}

%%%%%%%%%%%%%%%%%%%%%%%%%%%%%%%%%%%%%%%%%%%%%%%%%%%%%%%%%%%%%%%%%%
\subsection{Stratified partition} \label{subsec:partition}

As described in the previous section, the TxPI-u corpus contains essays from 416 subjects, each of them presents five traits according to the Big Five Model and every subject can be found in each trait (i.e. in Table \ref{table:sujetosporclase} the total number of subjects per trait is always 416). While this organization of TxPI-u can be useful to analyze traits independently to each other; might be a complication to analyze more than one trait in combination. Therefore, we decided to make a stratified partition where each trait has a set of representative examples from the positive (high) and negative (low) pole; as well as a set of examples for control purposes (control sample).

In the stratified partition of the TxPI-u, the control sample contains all the subjects with classes ``medium high'' or ``medium low'' for every trait. In other words, all subjects in the control group do not have any predominant traits (``high'' or ``low''). Hence, there is only one control group.

Subjects with representative traits, i.e. subjects labeled with classes ``high'' or ``low'' in only \textit{one} trait are selected for the stratified partition. This idea of stratified corpus has been done before by Oberlander and Gill, as they stated, a three-way stratified corpus allows to analyze features along a unique dimension \cite{oberlanderetal2006}.

Table \ref{table:partitiongral} shows the number of male and female per trait and the number of subjects in the control group of the resulting stratified partition. As we can see, the stratified corpus is smaller (almost half of the complete corpus), but it allows to perform a more fine analysis of the differences between traits. 

\begin{table}[tbp]
\centering
\caption{Number of subjects by gender and number of subjects by class, per trait, in the stratified partition} \label{table:partitiongral}
\begin{footnotesize}
\begin{tabular}{lrrrrr}
\hline
 & \multicolumn{2}{c}{Gender} & \multicolumn{2}{c}{Classes} \\
\cmidrule(lr){2-3} \cmidrule(lr){4-5}
Trait &	Male	&	Female	& High	& Low  & Total\\
\hline
    Openness	&	18	&	14	&  16 & 16  &	32	\\
    Consciousness	&	11	&	17	& 3	& 25  &   28	\\
    Extroversion	&	15	&	10	& 17 & 8 &   25	\\
    Agreeableness	&	12	&	20	& 10 & 22 &   32	\\
    Emot. Stability	&	7	&	7	&  6 & 8  &	14	\\
    Control	&	44	&	39	& -	& - & 83	\\
\hline
    Total    &	107	&	107	& 52 & 79 & 214	\\
\hline
\end{tabular}

\end{footnotesize}
\end{table}

\begin{figure}[tbp]
\includegraphics[width=1\textwidth]{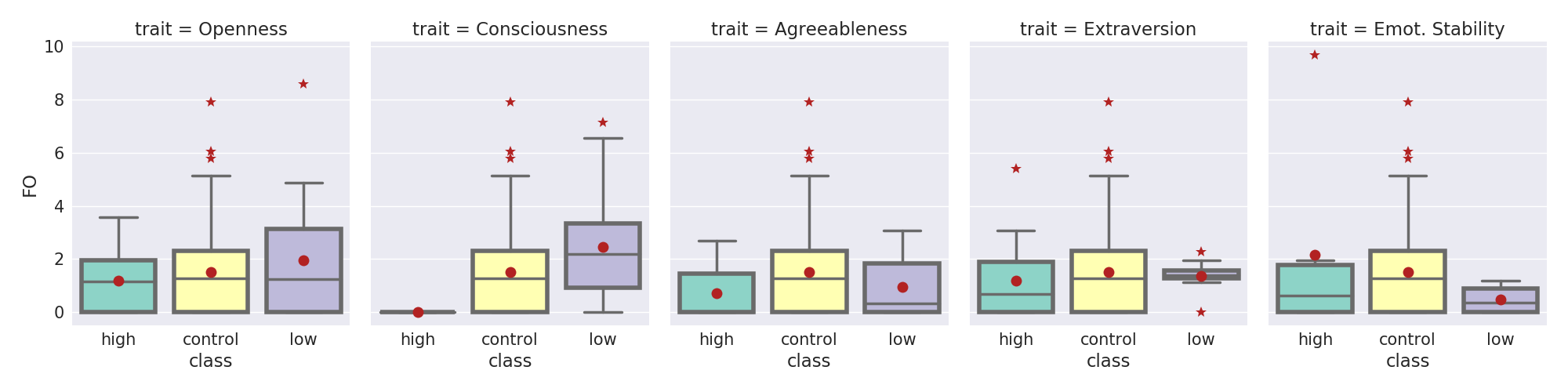}
\caption{Distribution of the misspelling tag (FO) across traits in the stratified corpus. The red dot inside the boxes shows the medium value of each sample and the dots beyond the limit of the boxes denote outliers.} \label{fig:foSTR}
\end{figure}

Regarding to handwritten phenomena, Table \ref{table:percentTagsSTR} shows the percentages of use of the seven tags. As can be noticed, the misspelling tag (FO) has the higher percentage. In Figure \ref{fig:foSTR} a closed look of the distribution of FO is presented.

\begin{table}[tbp]
\centering
\caption{Percentage of tags present in essays of the stratified partition of TxPI-u corpus. The tags FO, D, IN, MD, DL and NS means misspelling, drawing, insertion of some letter, modification of some word, elimination of some word, do not separate two words, and syllabification, respectively} \label{table:percentTagsSTR}
\begin{footnotesize}
\begin{tabular}{lrrrr}
\hline
Tag &	mean	&	std	& min	& max \\
\hline
    FO	& 1.46	&	1.71	&  0  & 9.68	\\
    D	& 0.01	&	0.14	& 0  &  1.92 \\
    IN	& 0.01	&	0.09	& 0 &  1.22	\\
    MD	& 0.99	&	1.99	& 0 &   20.64	\\
    DL	& 0.22  &	0.72	&  0 &	6.33	\\
    NS	& 0.42	&	0.96	& 0	& 6.04	\\
    SB  & 0.14	&	0.58	& 0 & 5.56	\\
\hline
\end{tabular}

\end{footnotesize}

\end{table}

In Figure \ref{fig:foSTR} is worth noticing the difference in the percentage of misspelling between classes of the same trait. For instance, for the trait extroversion there is almost the same percentage of misspelling for all subjects in the low class, while there is a bigger distribution of percentage of misspelling for subjects in the high class. That could be an indication of possible correlation of misspelling with a subject's predominant personality. 

%%%%%%%%%%%%%%%%%%%%%%%%%%%%%%%%%%%%%%%%%%%%%%%%%%%%%%%%%%%%%%%%%%
%%%%%%%%%%%%%%%%%%%%%%%%%%%%%%%%%%%%%%%%%%%%%%%%%%%%%%%%%%%%%%%%%%
%%%%%%%%%%%%%%%%%%%%%%%%%%%%%%%%%%%%%%%%%%%%%%%%%%%%%%%%%%%%%%%%%%

\section{Comparison of corpora for personality identification} \label{sec:comparison}

As it was mentioned in Section \ref{sec:relatedwork}, one corpus of English texts is the more related to TxPI-u. This corpus is a collection of essays compiled by Pennebaker and King \cite{Pennebakeretal99} during 1997 and 1999. Most of the texts from this corpus come from psychology students (approx. 1200). This initial corpus was increased by Mairesse et al. \cite{mairesse2007} with essays of students written until 2004. The final corpus, referred as \textit{Essays corpus} is a compilation of 2468 essays, most of them written by students. There is not information about the gender and age of participants in this corpus\footnote{In \cite{Pennebakeretal99} there is partial information about gender and age of 1200 participants approximately.}.

\begin{figure}[tbp]
\centering
\includegraphics[width=0.6\textwidth]{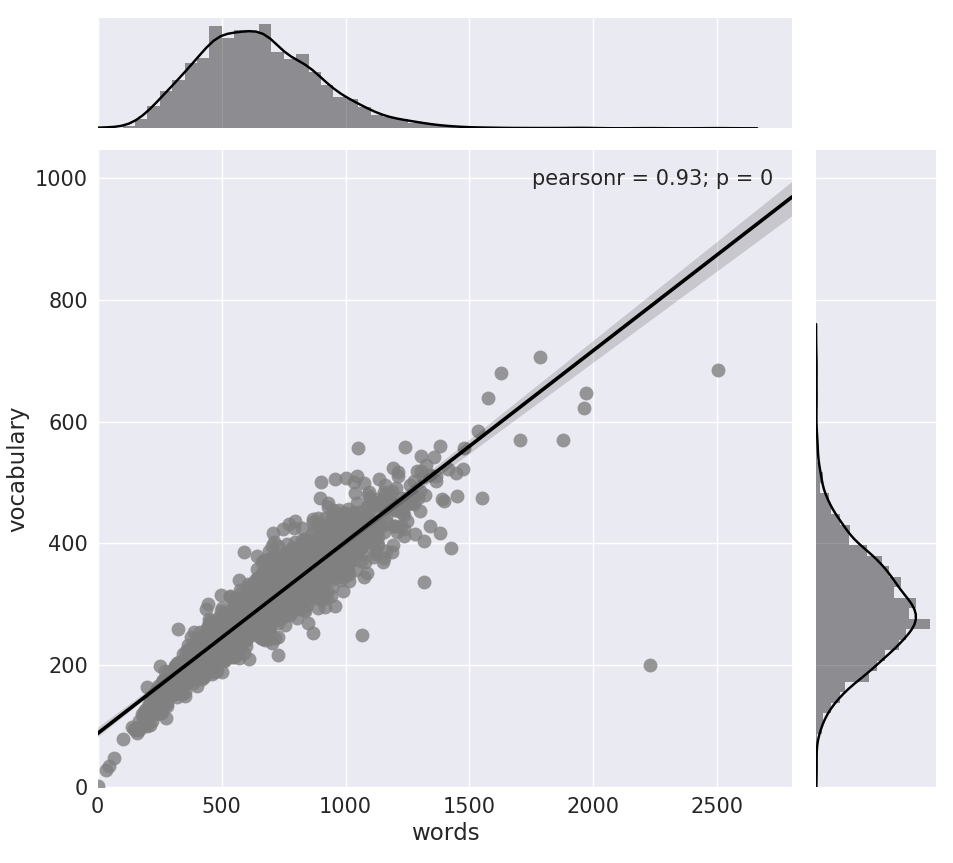}
\caption{Distribution of number of words and total vocabulary per essay and the correlation among these two variables in the Essays corpus. Frequency distribution graph (above and right) describe one variable independently of the other.}
 \label{fig:vocwordsEssays}
\end{figure}

\begin{figure}[tbp]
\centering
\includegraphics[width=0.6\textwidth]{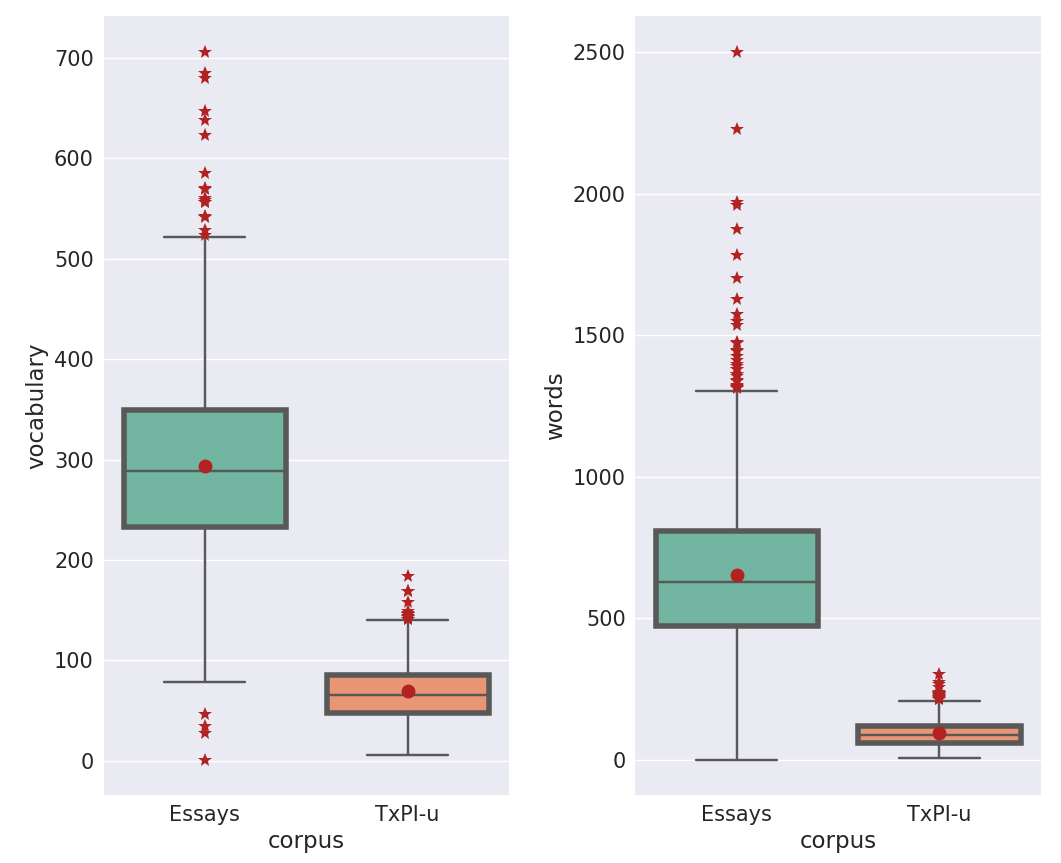}
\caption{Comparison of number of words and total vocabulary of each texts between the Essays corpus and the TxPI-u corpus. The red dot inside the boxes shows the medium value of each sample and the dots beyond the limit of the boxes denote outliers.} \label{fig:comparisonCorpus}
\end{figure}

In order to provide some general comparison between TxPI-u and the Essays corpus we show information about the number of words and vocabulary used (see Figure \ref{fig:vocwordsEssays}). Despite of the sample size (2468 vs 416 of Essays and TxPI-u respectively) there is a similar correlation between the number of words and the vocabulary used for each subject. In Figure \ref{fig:comparisonCorpus} a direct comparison between two variables is shown side by side. As can be seen, the texts in Essays are larger but also have more variation in the number of words used as in the vocabulary. This variation appears to a lesser extent in TxPI-u. The difference between the number of words used in both corpora could be explained in terms to the instruction given to the subjects. While in our case we asked for a personal experience, in the Essay corpus compilation, authors asked the subjects to write, for 20 uninterrupted minutes, anything that came to their minds (a complete description of the Essay corpora compilation can be found in \cite{Pennebakeretal99}).

Additionally, Table \ref{table:traitsEssays} shows the number of subjects per trait, note that there is not information about the numerical values of subjects' personality traits, only the nominal class ``yes'' or ``no'' was provided (the class ``yes'' is similar to our class ``high'' and the class ``no'' is similar to our class ``low''). With numerical values of personality's traits another normative values can be used to generate other partition of nominal classes or, a regression approach can be used to determined such values instead of just closed categories. 

\begin{table}[tbp]
\centering
\caption{Number of subjects per trait in the Essays corpus. Note that labels \textit{yes} and \textit{no} correspond to \textit{high} and \textit{low}, respectively} \label{table:traitsEssays}
\begin{footnotesize}

\begin{tabular}{lrr}
\hline
Trait &	 yes	& no \\
\hline
    Openness	&	1271	&	1196	\\
    Consciousness	&	1254	&	1214	\\
    Extroversion	&	1277	&	1191		\\
    Agreeableness	&	1310	&	1158	\\
    Emot. Stability	&	1235	&	1233	\\
\hline

\end{tabular}

\end{footnotesize}

\end{table}
%%%%%%%%%%%%%%%%%%%%%%%%%%%%%%%%%%%%%%%%%%%%%%%%%%%%%%%%%%%%%%%%%%
%%%%%%%%%%%%%%%%%%%%%%%%%%%%%%%%%%%%%%%%%%%%%%%%%%%%%%%%%%%%%%%%%%
%%%%%%%%%%%%%%%%%%%%%%%%%%%%%%%%%%%%%%%%%%%%%%%%%%%%%%%%%%%%%%%%%%

\section{Text classification with TxPI-u corpus: baselines} \label{sec:baselines}
The main goal of this section is to provide a set of baselines for the text classification task of personality identification. All the experiments reported in this section were done using the stratified partition of the TxPI-u corpus. 

We provide a set of basic configuration systems, widely employed in the text classification (TC) task. Obtained results will serve for comparison purposes against future methods or future representations. The intention is to use representations such as n-grams of words, n-grams of characters and n-grams of part of speech (POS) in combination with the most common learning algorithms for text classification such as naive bayes, decision trees and support vector machines.

%%%%%%%%%%%%%%%%%%%%%%%%%%%%%%%%%%%%%%%%%%%%%%%%%%%%%%%%%%%%%%%%%%
\subsection{Evaluation metrics} \label{subsec:evalmetric}
The evaluation metric used was the macro-averaged $F_{1}$, also known as F-score. This measure allows to obtain confident perspectives of the system's performance, particularly for cases where classes are highly unbalanced, such as in the stratified partition of TxPI-u.
%%%%%%%%%%%%%%%%%%%%%%%%%%%%%%%%%%%%%%%%%%%%%%%%%%%%%%%%%%%%%%%%%%

\subsection{Experimental setup} \label{subsec:setup}

Five classification problems were defined, one per trait. Each problem has three classes: high, low, and control. We represented each essay using three different type of representations: n-grams of words, n-grams of characters and n-grams of POS (Part of Speech) tags. For each type we used n-grams' sizes of 1, 2 and 3 for words and POS, and n-grams' sizes of 3, 4, 5 for characters. In addition, for each experiment we used three different classifiers: Naive Bayes, J48 and SMO\footnote{For all experiments we used Weka Tool Kit \cite{Nicholas1995} with the default configuration values.}. 

A vector space model was used to represent each text; thus, for each essay we have a multi-dimensional vector. In this vector, we evaluated three different weighing schemes: boolean (the importance of each term in the vector should be 1 if the term appears in the document, and 0 otherwise), term frequency (the number of times a term appears in the document), and tf-idf (the importance of a term given by the term frequency and the inverse document frequency).

%%%%%%%%%%%%%%%%%%%%%%%%%%%%%%%%%%%%%%%%%%%%%%%%%%%%%%%%%%%%%%%%%%
\subsection{Results} \label{subsec:results}

Altogether, we performed 405 experiments (5 traits, 9 representations, 3 weights schemes, and 3 learning algorithms) and for all of them we used 10 fold cross validation to evaluate each three-class classifier. For all results we calculate the F-score that can be seen in Table \ref{table:results}. Note that in the results' table only boolean (bool) and term frequency (tf) as weighting schema is shown, that is because tf-idf (term frequency - inverse document frequency) performs similar to tf, therefore, we only show tf results.

Additionally, a set of experiments were done using a Bag of Words representation with the words categories presented in the Spanish dictionary (version 2007) of LIWC (Linguistic Inquiry and Word Count) \cite{Pennebaker2007}. The results of this experiment also reported using the F-score metric; see Table \ref{table:resultsLIWC}.

\begin{table}[tbp]
\centering
\caption{Classification results using ten cross fold validation technique with the stratified corpus of TxPI-u. Each result corresponds to a three class classification problem. Results are given in F-score and the representation used is BOW with LIWC tags} 
\label{table:resultsLIWC}
\begin{footnotesize}
\begin{tabular}{rrrrrrr}
\hline
& \multicolumn{2}{c}{NB} & \multicolumn{2}{c}{SMO} & \multicolumn{2}{c}{J48} \\
\cmidrule(lr){2-3} \cmidrule(lr){4-5} \cmidrule(lr){6-7} 

	&	bool	&	tf	&	bool	&	tf	&	bool	&	tf	\\
Ope	&	0.31	&	0.33	&	\textbf{0.35}	&	0.26	&	0.34	&	0.34	\\
Con	&	0.30	&	\textbf{0.36}	&	0.29	&	0.33	&	0.26	&	\textbf{0.36}	\\
Ext	&	0.31	&	0.34	&\textbf{	0.35}	&	0.34	&	0.34	&	0.33	\\
Agr	&	0.34	&	0.40	&	0.31	&	0.27	&	\textbf{0.43}	&	0.36	\\
Sta	&	0.35	&	\textbf{0.40}	&	0.27	&	0.30	&	0.31	&	0.37	\\
\hline
\end{tabular}

\end{footnotesize}

\end{table}

Overall, the performance of a three-class problem is not superior to 0.49 of F-score which illustrates the difficulty of this problem. We believe that by means of novel representation schemes or new learning methods, obtained results can be significantly improved. 

Despite the low performances, we can have some interesting insights about the representations used. For instance, the representation based on POS tags is the best across three traits: Openness (f-score of 0.49 with uni-grams), Consciousness (f-score of 0.39 with bi-grams) and Emotional Stability (f-score of 0.40 with uni-grams); while 5-grams of characters performed better for Extroversion and Agreeableness (0.45 of f-score for both cases). 

Regarding the representation using word categories of LIWC, the performance of classification was not better that using an open-vocabulary approach (as in the experiments reported in Table \ref{table:results}). It is clear that these results are not definitive, and there still is an important room for improvement, e.g. a representation that combine open-vocabulary with word categories. Even more, a deeper analysis can be performed to better understand the difficulty of the personality identification problem.

\begin{table*}[tbp]
\centering
\caption{Classification results using a ten fold cross validation technique with the stratified corpus of TxPI-u. Each result corresponds to a three-class classification problem. Results are given in F-score} \label{table:results}
\begin{tiny}
\begin{tabular}{rrrrrrrrrrrrrrrrrrr}
\hline
\multicolumn{19}{c}{Words}\\
 & \multicolumn{6}{c}{1-gram} &  \multicolumn{6}{c}{2-gram} & \multicolumn{6}{c}{3-gram}\\
\cmidrule(lr){2-7} \cmidrule(lr){8-13} \cmidrule(lr){14-19} 
& \multicolumn{2}{c}{NB} & \multicolumn{2}{c}{SMO} & \multicolumn{2}{c}{J48} & \multicolumn{2}{c}{NB} & \multicolumn{2}{c}{SMO} & \multicolumn{2}{c}{J48} & \multicolumn{2}{c}{NB} & \multicolumn{2}{c}{SMO} & \multicolumn{2}{c}{J48}\\
\cmidrule(lr){2-3} \cmidrule(lr){4-5} \cmidrule(lr){6-7}  \cmidrule(lr){8-9} \cmidrule(lr){10-11} \cmidrule(lr){12-13}  \cmidrule(lr){14-15} \cmidrule(lr){16-17} \cmidrule(lr){18-19} 
&	bool	&	tf	&	bool	&	tf	&	bool	&	tf	&	bool	&	tf	&	bool	&	tf	&	bool	&	tf	&	bool	&	tf	&	bool	&	tf	&	bool	&	tf	\\
\hline
Ope	&	0.28	&	0.34	&	0.28	&	0.28	&	0.35	&	0.36	&	0.28	&	0.33	&	0.28	&	0.28	&	\textbf{0.38}	&	0.32	&	0.28	&	0.27	&	0.28	&	0.28	&	0.28	&	0.27	\\
Con	&	0.27	&	0.32	&	0.30	&	0.28	&	0.30	&	0.36	&	0.27	&	0.35	&	0.28	&	0.28	&	\textbf{0.38}	&	0.32	&	0.29	&	0.33	&	0.29	&	0.29	&	0.28	&	0.32	\\
Ext	&	0.29	&	0.31	&	0.29	&	0.29	&	\textbf{0.36}	&	\textbf{0.36}	&	0.29	&	0.27	&	0.29	&	0.29	&	0.26	&	\textbf{0.36}	&	0.29	&	0.34	&	0.29	&	0.29	&	0.29	&	0.29	\\
Agr	&	0.35	&	\textbf{0.45}	&	0.35	&	0.31	&	0.31	&	0.35	&	0.28	&	0.26	&	0.28	&	0.28	&	0.30	&	0.33	&	0.28	&	0.24	&	0.28	&	0.28	&	0.27	&	0.27	\\
Sta	&	0.31	&	0.39	&	0.31	&	0.31	&	0.35	&	0.39	&	0.31	&	0.36	&	0.31	&	0.31	&	0.31	&	\textbf{0.41}	&	0.31	&	0.37	&	0.31	&	0.31	&	0.31	&	0.31	\\
\hline
\multicolumn{19}{c}{Part of Speech} \\
 & \multicolumn{6}{c}{1-gram} &  \multicolumn{6}{c}{2-gram} & \multicolumn{6}{c}{3-gram}  \\
\cmidrule(lr){2-7} \cmidrule(lr){8-13} \cmidrule(lr){14-19} 
& \multicolumn{2}{c}{NB} & \multicolumn{2}{c}{SMO} & \multicolumn{2}{c}{J48} & \multicolumn{2}{c}{NB} & \multicolumn{2}{c}{SMO} & \multicolumn{2}{c}{J48} & \multicolumn{2}{c}{NB} & \multicolumn{2}{c}{SMO} & \multicolumn{2}{c}{J48}  \\
\cmidrule(lr){2-3} \cmidrule(lr){4-5} \cmidrule(lr){6-7}  \cmidrule(lr){8-9} \cmidrule(lr){10-11} \cmidrule(lr){12-13}  \cmidrule(lr){14-15} \cmidrule(lr){16-17} \cmidrule(lr){18-19} 
	&	bool	&	tf	&	bool	&	tf	&	bool	&	tf	&	bool	&	tf	&	bool	&	tf	&	bool	&	tf	&	bool	&	tf	&	bool	&	tf	&	bool	&	tf	\\
Ope	&	0.41	&	0.40	&	0.37	&	\textbf{0.49}	&	0.38	&	0.42	&	0.29	&	0.30	&	0.30	&	0.30	&	0.34	&	0.44	&	0.32	&	0.34	&	0.28	&	0.28	&	0.24	&	0.28	\\
Con	&	0.36	&	0.35	&	0.42	&	0.34	&	0.31	&	0.30	&	0.28	&	\textbf{0.39}	&	0.30	&	0.33	&	0.28	&	0.33	&	0.30	&	0.37	&	0.29	&	0.28	&	0.29	&	0.27	\\
Ext	&	\textbf{0.39}	&	0.32	&	0.37	&	0.28	&	0.28	&	0.34	&	0.28	&	0.28	&	0.35	&	0.31	&	0.30	&	0.27	&	0.29	&	0.27	&	0.29	&	0.29	&	0.31	&	0.25	\\
Agr	&	0.34	&	0.33	&	\textbf{0.41}	&	0.37	&	0.29	&	0.32	&	0.34	&	0.29	&	0.33	&	0.29	&	0.31	&	0.30	&	0.34	&	0.29	&	0.28	&	0.28	&	0.26	&	0.29	\\
Sta	&	0.33	&	0.28	&	\textbf{0.46}	&	0.31	&	0.31	&	0.29	&	0.31	&	0.31	&	0.38	&	0.31	&	0.41	&	0.41	&	0.31	&	0.31	&	0.31	&	0.31	&	0.37	&	0.29	\\
\hline
\multicolumn{19}{c}{Characters} \\
 & \multicolumn{6}{c}{3-gram} &  \multicolumn{6}{c}{4-gram} & \multicolumn{6}{c}{5-gram}  \\
\cmidrule(lr){2-7} \cmidrule(lr){8-13} \cmidrule(lr){14-19} 
& \multicolumn{2}{c}{NB} & \multicolumn{2}{c}{SMO} & \multicolumn{2}{c}{J48} & \multicolumn{2}{c}{NB} & \multicolumn{2}{c}{SMO} & \multicolumn{2}{c}{J48} & \multicolumn{2}{c}{NB} & \multicolumn{2}{c}{SMO} & \multicolumn{2}{c}{J48}  \\
\cmidrule(lr){2-3} \cmidrule(lr){4-5} \cmidrule(lr){6-7}  \cmidrule(lr){8-9} \cmidrule(lr){10-11} \cmidrule(lr){12-13}  \cmidrule(lr){14-15} \cmidrule(lr){16-17} \cmidrule(lr){18-19} 
	&	bool	&	tf	&	bool	&	tf	&	bool	&	tf	&	bool	&	tf	&	bool	&	tf	&	bool	&	tf	&	bool	&	tf	&	bool	&	tf	&	bool	&	tf	\\
Ope	&	0.31	&	0.28	&	0.28	&	0.27	&	0.37	&	0.26	&	0.28	&	0.28	&	0.28	&	0.28	&	0.28	&	0.33	&	0.27	&	0.27	&	0.28	&	0.28	&	\textbf{0.38}	&	\textbf{0.38}	\\
Con	&	0.34	&	0.28	&	0.30	&	0.31	&	0.37	&	0.36	&	0.35	&	0.28	&	0.34	&	0.30	&	\textbf{0.38}	&	0.33	&	0.32	&	0.33	&	0.31	&	0.31	&	0.29	&	0.31	\\
Ext	&	0.35	&	0.29	&	0.36	&	0.33	&	0.33	&	0.33	&	0.32	&	0.28	&	0.29	&	0.29	&	0.30	&	0.29	&	0.32	&	0.29	&	0.29	&	0.29	&	\textbf{0.45}	&	0.41	\\
Agr	&	0.36	&	0.27	&	0.35	&	0.41	&	0.38	&	0.35	&	0.37	&	0.28	&	0.33	&	0.33	&	0.26	&	0.37	&	0.39	&	0.37	&	0.28	&	0.28	&	\textbf{0.45}	&	0.43	\\
Sta	&	0.31	&	0.31	&	0.31	&	0.31	&	0.29	&	0.35	&	0.31	&	0.31	&	0.31	&	0.31	&	0.27	&	\textbf{0.44}	&	0.31	&	0.31	&	0.31	&	0.31	&	0.31	&	0.30	\\
\hline
\end{tabular}
\end{tiny}
\end{table*}

%%%%%%%%%%%%%%%%%%%%%%%%%%%%%%%%%%%%%%%%%%%%%%%%%%%%%%%%%%%%%%%%%%
%%%%%%%%%%%%%%%%%%%%%%%%%%%%%%%%%%%%%%%%%%%%%%%%%%%%%%%%%%%%%%%%%%
%%%%%%%%%%%%%%%%%%%%%%%%%%%%%%%%%%%%%%%%%%%%%%%%%%%%%%%%%%%%%%%%%%
\section{Conclusions} \label{sec:conclusions}
In this paper we introduced a corpus of Spanish texts annotated with personality information, age and gender for each author of those texts. The corpus, named TxPI-u, Text for Personality Identification of undergraduate, is a collection of 416 short essays of undergraduate Mexican students. We have described TxPI-u corpus in terms of number of words, vocabulary and distribution frequencies among personality traits, particularly within the Big Five personality model.

From the texts in the 416 essays, we manually labelled some handwriting phenomena, such as, modification of a word, insertion of some letter, the use of emojis or drawings, etc. We also labelled all the misspelling words found in those essays as well as syllabification acts. In this direction, we found that there is no clear correlation among the seven labels used. Nevertheless, this manually labelling of handwritten phenomena is not found, to the best of our knowledge, in any other corpus for personality identification. A work perspective is a correlation study of the presence of each label and a personality trait, one intuition is that misspelling can be associated with Consciousness in the low class.

In order to allow a fine analysis of personality traits, we created a stratified partition from TxPI-u corpus. The resulting partition contains a total of 214 subjects. In this direction, a work perspective is to study a binary classification problem, with one class for the pole of interest (either high or low) and the other class with only the control class. In this configuration only important markers for a class of a given trait will be analyzed in isolation, thus correlations can be found without introducing noise. A different, yet interesting line of work, would be to face the personality identification problem as a regression problem, i.e., identify the numeric values of each trait.  

As an additional contribution of this work, we describe a the set of baselines to provide a comparison point for further research in author profiling, specifically for personality identification. For these experiments, we used as main form of representation n-grams of: words, characters and part of speech, as well as, word categories provided by the LIWC Spanish dictionary. Tackling both, closed-vocabulary approach and open-vocabulary approach.

Finally, we believe that this resource, carefully gathered, represents an important contribution to the community doing research in the area of personality identification, and in general for the author profiling research field. With the extra information attached to the essays, automatic models can be proposed for identifying gender, age, and academic program election. In addition, our built corpus has a multimodal applicability, since it could be interesting to analyze the handwritten phenomena using novel computer vision strategies. All these characteristics make the TxPIu corpus a more challenging dataset. The complete corpus as well as the instrument used for collected it can be found at \url{lyr.cua.uam.mx/resources/personality/TxPIu/}. 

\section*{Acknowledgments}
We would like to thank to the Universidad Autónoma Metropolitana Unidad Cuajimalpa (UAM-C) for all the facilities provided for developing this research project. Particularly, we thank to Eduardo Peñalosa-Castro and Liliana Dávalos-Alcázar.  Furthermore, this work was partially supported by CONACYT under project grant 258588 and under the Thematic Networks program (Language Technologies Thematic Network, project 281795).

%%%%%%%%%%%%%%%%%%%%%%%%%%%%%%%%%%%%%%%%%%%%%%%%%%%%%%%%%%%%%%%%%%
%%%%%%%%%%%%%%%%%%%%%%%%%%%%%%%%%%%%%%%%%%%%%%%%%%%%%%%%%%%%%%%%%%
%%%%%%%%%%%%%%%%%%%%%%%%%%%%%%%%%%%%%%%%%%%%%%%%%%%%%%%%%%%%%%%%%%

\bibliographystyle{plain}
\bibliography{main.bib}

\end{document}